\documentclass[sigconf]{acmart}
\usepackage{multirow}
\usepackage{amsmath}
\usepackage{algorithm}
\usepackage[noend]{algpseudocode}

\makeatletter
\newcounter{phase}[algorithm]
\newlength{\phaserulewidth}
\newcommand{\setphaserulewidth}{\setlength{\phaserulewidth}}
\newcommand{\phase}[1]{%
  \vspace{-1.25ex}
  \Statex\leavevmode\llap{\rule{\dimexpr\labelwidth+\labelsep}{\phaserulewidth}}\rule{\linewidth}{\phaserulewidth}
  \Statex\strut\refstepcounter{phase}\textit{Stage~\thephase~--~#1}% Phase text
  \vspace{-1.25ex}\Statex\leavevmode\llap{\rule{\dimexpr\labelwidth+\labelsep}{\phaserulewidth}}\rule{\linewidth}{\phaserulewidth}}
\makeatother
\setphaserulewidth{.7pt}

\algnewcommand\algorithmicinput{\textbf{Input:}}
\algnewcommand\algorithmicoutput{\textbf{Output:}}
\algnewcommand\Input{\item[\algorithmicinput]}%
\algnewcommand\Output{\item[\algorithmicoutput]}%
\AtBeginDocument{%
  \providecommand\BibTeX{{%
    \normalfont B\kern-0.5em{\scshape i\kern-0.25em b}\kern-0.8em\TeX}}}

\copyrightyear{2021}
\acmYear{2021}
\setcopyright{acmcopyright}
\acmConference[MM '21] {Proceedings of the 29th ACM International Conference on Multimedia}{October 20--24, 2021}{Virtual Event, China.}
\acmBooktitle{Proceedings of the 29th ACM Int'l Conference on Multimedia (MM '21), Oct. 20--24, 2021, Virtual Event, China}
\acmPrice{15.00}
\acmDOI{10.1145/3474085.3475663}  %updated DOI
\acmISBN{978-1-4503-8651-7/21/10}
\begin{document}
\fancyhead{}

\title{Unsupervised Portrait Shadow Removal via Generative Priors}

\author{Yingqing He}
\authornote{Both authors contributed equally.}
\affiliation{%
  \institution{HKUST}
%   \city{Hong Kong} 
  \country{ }
}
\author{Yazhou Xing}
\authornotemark[1]
\affiliation{%
  \institution{HKUST}
%   \city{Hong Kong} 
\country{ }
}
% \author{}
% \affiliation{%
%   \institution{The Hong Kong University of Science and Technology}
% %   \city{Hong Kong} \country{China}
%   \city{} \country{}}
\author{Tianjia Zhang}
\affiliation{%
  \institution{HKUST}
%   \city{Hong Kong} 
\country{ }
}
\author{Qifeng Chen}
\affiliation{%
  \institution{HKUST}
%   \city{Hong Kong} 
\country{ }
}

% \maketitle

% \author{Yingqing He}
% \affiliation{%
%   \institution{}
% %   \city{Hong Kong} \country{China}
%   \city{} \country{}
% }

% \author{Yingqing He }
% \affiliation{%
%   \institution{}
% %   \city{Hong Kong} \country{China}
%   \city{} \country{}
% }
% \author{Yingqing He }
% \affiliation{%
%   \institution{}
% %   \city{Hong Kong} \country{China}
%   \city{} \country{}
% }
% \author{Yingqing He }
% \affiliation{%
%   \institution{}
% %   \city{Hong Kong} \country{China}
%   \city{} \country{}
% }

% \maketitle
\begin{abstract}
Portrait images often suffer from undesirable shadows cast by casual objects or even the face itself. While existing methods for portrait shadow removal require training on a large-scale synthetic dataset, we propose the first unsupervised method for portrait shadow removal without any training data. Our key idea is to leverage the generative facial priors embedded in the off-the-shelf pretrained StyleGAN2. To achieve this, we formulate the shadow removal task as a layer decomposition problem: a shadowed portrait image is constructed by the blending of a shadow image and a shadow-free image. We propose an effective progressive optimization algorithm to learn the decomposition process. Our approach can also be extended to portrait tattoo removal and watermark removal. Qualitative and quantitative experiments on a real-world portrait shadow dataset demonstrate that our approach achieves comparable performance with supervised shadow removal methods. Our source code is available at \href{https://github.com/YingqingHe/Shadow-Removal-via-Generative-Priors}{\color{blue}this repository}.
% \url{ https://github.com/YingqingHe/Shadow-Removal-via-Generative-Priors}.
% , while our method does not require a large training dataset
\end{abstract}

\ccsdesc[500]{Computing methodologies~Computational photography}
\ccsdesc[500]{Computing methodologies~Image processing}

\vspace{-3mm}
\keywords{Portrait Shadow Removal, Unsupervised Learning, Generative Priors, Image Decomposition}

\begin{teaserfigure}
    \begin{tabular}{c@{\hspace{0.7mm}}c@{\hspace{0.7mm}}c@{\hspace{0.7mm}}c@{\hspace{0.7mm}}c@{\hspace{0.7mm}}c@{}}
    \rotatebox{90}{\hspace{15mm} Input}&
    \includegraphics[width=0.24\linewidth]{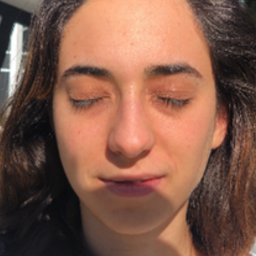}&
    \includegraphics[width=0.24\linewidth]{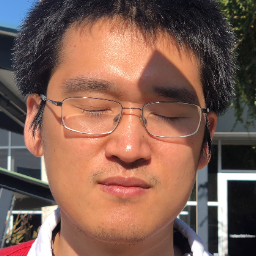}&
    \includegraphics[width=0.24\linewidth]{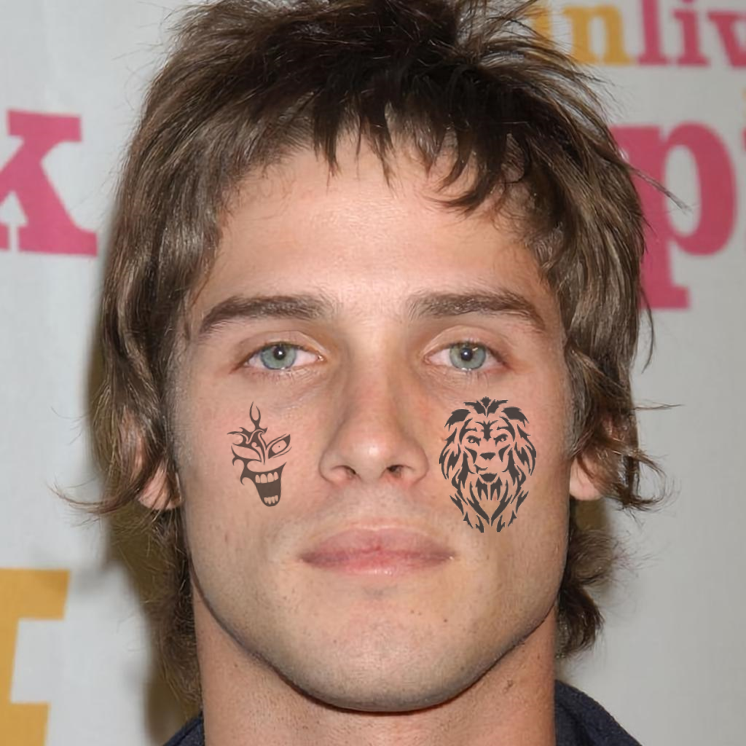}&
    \includegraphics[width=0.24\linewidth]{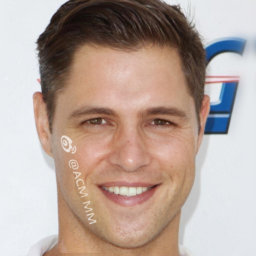}\\
    \rotatebox{90}{ \hspace{11mm} Our results}&
    \includegraphics[width=0.24\linewidth]{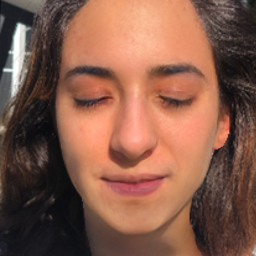}&
    \includegraphics[width=0.24\linewidth]{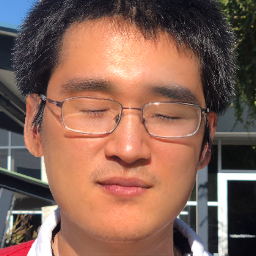}&
    \includegraphics[width=0.24\linewidth]{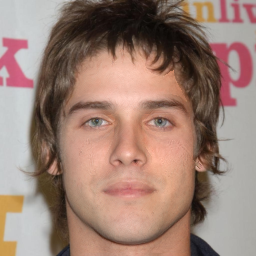}&
    \includegraphics[width=0.24\linewidth]{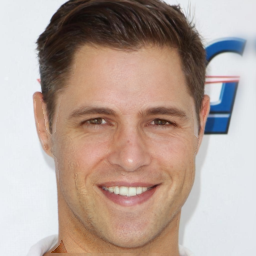}\\
    &Shadow removal&Shadow removal&Tattoo removal &Watermark removal \\
    &&&(extension)&(extension)
    \end{tabular}
    \vspace{-2mm}
    \caption{\textbf{Our portrait shadow removal, tattoo removal, and watermark removal results.} We propose the first unsupervised portrait shadow removal method without any training data. Our method can recover high-quality shadow-free portrait images from real-world portrait shadow images. Our proposed method can also be extended to facial tattoo removal and watermark removal as a general framework with only little modification. }
  \label{fig:teaser}
\end{teaserfigure}

\maketitle

\section{Introduction}
% What is the problem? Why is it important? 
%% portrait shadow removal and tatto/watermark removal. good for photography. 
Portrait photography is the art of capturing the inherent character of a person with a face in a photograph. Due to the rapid growth of digital cameras and smartphone cameras, portrait photography has also become popular among amateur photographers. However, portrait photographs taken in the wild often suffer from undesirable shadows cast by casual objects or even the face itself due to the lack of professional lighting control or the unpleasing environmental illumination conditions. Although photo editing software such as Adobe Photoshop provides a series of image adjustment operations for post-processing, portrait shadow removal as an effective, high-quality, and automatic application is highly desirable. Therefore, we are interested in designing an automatic and effective algorithm to remove portrait shadows without any user input.

% Why is the problem hard? What makes it challenging? 
%% lack of real world datasets. problem itself is hard. 
%% face requires high quality makes the problem challenging, but we have stylegan. 

% How far has exiting works come? Why hasn’t the problem been solved? What is the stumbling block? What is the frontier?
%% existing portrait shadow removal: rely on large-scale training data, may perform not good on real-world scenes; categorize existing work into 2-3 classes 
% Cun et al. / layer decomposition / sig
Due to the importance of lighting manipulation to photography, there have been many approaches to remove unpleasing shadows in photographs~\cite{wu2007natural,shor2008shadow, arbel2010shadow,gryka2015learning,drew2003recovery, finlayson20014, finlayson2005removal, xiao2013fast, guo2012paired, wu2007natural,qu2017deshadownet, wang2018stacked, hu2018direction, le2019shadow, cun2020towards}. Recent state-of-the-art shadow removal and portrait shadow removal methods are based on deep learning techniques and trained on large-scale image pairs in a supervised manner~\cite{qu2017deshadownet, wang2018stacked, hu2018direction, le2019shadow, cun2020towards}. Although it has achieved better performance than traditional methods, those approaches are limited by the nature of their training data which may fail on complex and various real-world images. Moreover, preparing training data is a tedious and challenging task, since the diversity of training pairs for varied background environments, shadows, and subjects are very hard to cover. Meanwhile, different tasks need different kinds of training data also causes inconvenience on both model training and practical usage.

In this paper, we propose the first unsupervised portrait shadow removal method without any paired training data.
Our method takes only one single shadowed portrait image as input, then recovers the shadow-free portrait image without any user intervention. The key insight of our method is that off-the-shelf pretrained face generators such as StyleGAN2~\cite{karras2020analyzing}, own abundant priors on high-quality face appearances and geometries. The prior work~\cite{pan2020exploiting} also exploits generative priors for versatile image restoration and manipulation tasks 
including colorization and inpainting. However, their method assumes that the degradation transformation is deterministic (e.g., converting generated RGB image to grayscale for image colorization and a known binary mask for image inpainting). This assumption cannot hold for restoration cases where the image degradation is complex and unknown. In our portrait shadow removal task, the shadow degradation is spatially varying and the shadowed region can be any kinds of shape. Thus, unlike their method, we make the first attempt to exploiting generative priors towards unknown degradation process in order to solve the portrait shadow removal task in an unsupervised manner. 
To effectively leverage generative priors with unknown shadowed degradation, we formulate the portrait shadow removal task as an image decomposition problem~\cite{zhang2020portrait}: a shadowed image is a composite of a full-shadow image and a shadow-free image.Since learning the image layer decomposition from one single image is an ill-posed problem, we observed that exploiting deep generative facial priors can help to reduce the ambiguity of the layer decomposition. For the shadow degradation learning, we exploit a neural network which takes a random noise as input to estimate the shadow mask. Meanwhile, we optimize a color matrix to obtain a full shadow image from the generated shadow-free image. We note that directly optimizing our framework leads to poor performance due to the layer ambiguity nature of this problem. Thus, we propose a progressive optimization strategy and design effective regularization to clear the ambiguity step-by-step. Extensive experiments show that our unsupervised method can achieve comparable performance with state-of-the-art supervised methods. Moreover, we also extend our method to tattoo removal and watermark removal for portrait images to demonstrate the general ability of our method.

Our contributions can be summarized as follows:
% We use the pretrained StyleGAN to learn the underlying shadow-free image and optimize a color matrix to obtain the full shadow image. Inspired by DIP~\cite{ulyanov2018deep}, we use a neural network which takes a random noise as input to learn the mask for the blending of shadow-free image and full-shadow image. 

\begin{itemize}
\item We present the first unsupervised portrait shadow removal method leveraging the abundant deep generative priors embedded in the pretrained GAN model. Our model can handle unknown degradation process in shadowed images, which cannot be achieved by a set of existing GAN-inversion-based image restoration methods. 
\item We propose an effective progressive optimization strategy to eliminate the ambiguous nature of the shadow degradation learning from a single input image. 
% We also design specific regularization terms and loss functions to better empower our method to learn the shadow degradation and clean image decomposition.

%eliminate the ambiguity nature of portrait shadow removal step-by-step

\item We demonstrate that our unsupervised method can achieve comparable performance with existing supervised methods. In addition, our method can also serve as a general framework to deal with multiple tasks (i.e., watermark removal and tattoo removal), which cannot be achieved via existing supervised methods.

% \item We extend our method to tattoo removal and watermark removal to show our method can serve as a general framework for other image recovering tasks. 

% \item We conduct extensive experiments on a real-world portrait shadow dataset to demonstrate that our unsupervised method can achieve comparable performance without any training data, meanwhile owns better versatility than any existing supervised methods. 
\end{itemize}

\begin{figure*}[t]
\centering
\includegraphics[width=1.0\textwidth]{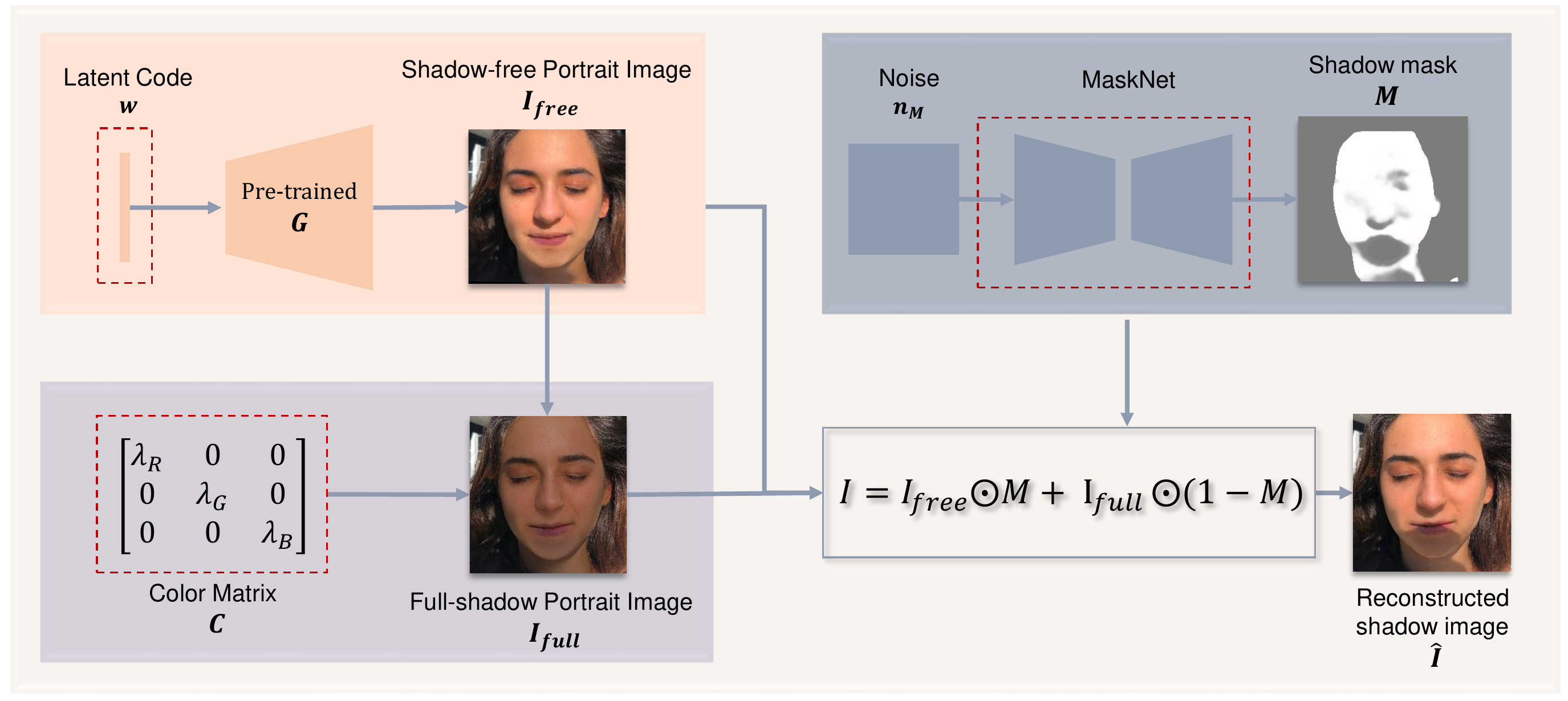}
\vspace{-0.7em}
\caption{\textbf{Framework Overview}. We design a multi-path network to effectively decompose a shadowed portrait image into three components: a shadow-free image $I_{free}$, a full-shadow image $I_{full}$, and a shadow mask $M$. We utilize generative priors from the pretrained StyleGAN to generate $I_{free}$ and optimize a color matrix $C$ to generate $I_{full}$. We propose a MaskNet to learn a shadow mask from sampled noise. Composing $I_{full}$ with $I_{free}$ with $M$ can reconstruct the input shadowed image $\hat{I}$(see Eq. ~\ref{formula:decomposition}). Blocks in red dash lines are trainable. Please refer to Section~\ref{method} for our detailed design and optimization strategy.}
\label{fig:overview}
\vspace{-0.7em}
\end{figure*}

\section{Related Work}
% shadow removal 
% Generative prior 
\subsection{Shadow removal}
Traditionally, graphics-based shadow removal approaches identify shadows by manually label shadow regions~\cite{wu2007natural,shor2008shadow, arbel2010shadow,gryka2015learning}, exploit shadow priors such as illumination discontinuity across shadow edges~\cite{sato2003illumination, baba2003shadow}, or find the relationship between shadow and non-shadow regions~\cite{guo2012paired}. Then, shadow removal can be performed by histogram manipulation~\cite{kaufman2012content}, color transfer~\cite{shor2008shadow, wu2005bayesian, vicente2017leave}, illumination modelling and relighting~\cite{drew2003recovery, finlayson20014, finlayson2005removal, xiao2013fast, guo2012paired, wu2007natural}.

With the development of deep learning in recent years, many works have been proposed to train deep neural networks on large-scale datasets for shadow removal~\cite{qu2017deshadownet, wang2018stacked, hu2018direction, le2019shadow, cun2020towards}. Deshadow-Net~\cite{qu2017deshadownet} removes shadows in an end-to-end manner to predict a shadow matte. ST-CGAN~\cite{wang2018stacked} exploits conditional GAN~\cite{isola2017image} to generate shadow-free images and shadow masks in a unified framework. Mask-ShadowGAN~\cite{hu2019mask} and ARGAN~\cite{ding2019argan} also exploit generative models~\cite{chen2016infogan} to perform shadow removal on unpaired training data or in a semi-supervised manner. Recently~\cite{le2019shadow} formulate the shadow removal task as an image layer decomposition problem similar to ours. 

However, all of the previous deep learning approaches need a large amount of paired images (e.g., shadow and shadow-free image pairs) to train their networks which is time-consuming. Besides, collecting training data pairs is tedious and collected training pairs may have inconsistent colors, luminosity, and positions~\cite{hu2019mask}. Most importantly, their method has dubious generalization ability due to the intrinsic limitation caused by their training data which may not be generalized well on portrait images.

For portrait shadow removal, ~\cite{zhang2020portrait} divide facial shadows into two types: foreign shadows which are cast by foreign objects, and facial shadows which are caused by the face itself. They designed two separate models for each kind of shadow, and two training sets are needed for these two tasks. While our upsupervised portrait shadow removal method can handle the two types of shadow in one single model and only use one single input shadowed portrait image.

\subsection{Deep generative priors}
Since performing shadow and face decomposition in a single-image manner is an ill-pose problem, we exploit the idea of GAN inversion~\cite{pan2020exploiting, menon2020pulse, zhu2020domain, gu2020image} so that we can use a pretrained state-of-the-art GAN model to provide high-quality generative facial priors to guide our clean portrait image reconstruction process. 

GAN inversion aims at finding a corresponding latent vector to reconstruct the desired image using a pretrained GAN generator. Thanks to the development of state-of-the-art GAN models, many works proposed to exploit the generative priors to facilitate a set of image restoration and processing tasks. Sachit et al.~\cite{menon2020pulse} proposed PULSE which aimed to find a corresponding latent code in the latent space of generative models to generate high-resolution images from a single low-resolution image. Gu et al.~\cite{gu2020image} increased the amount of latent code and expand the application of GAN priors to multiple image processing and manipulation tasks. Pan et al.~\cite{pan2020exploiting} proposed DGP to exploit the deep generative priors from the pretrained BigGAN to facilitate a set of image restoration tasks. 

However, previous works exploiting GAN priors always assume the image degradation process is known, while our method can learn the spatial-varying and various shapes of shadow degradation from the input single shadowed image, which is a more challenging setting than previous methods.

% \begin{figure*}[t]
% \centering
% \begin{tabular}{c@{\hspace{0.5mm}}c@{\hspace{0.5mm}}c@{\hspace{0.5mm}}c@{\hspace{0.5mm}}c@{\hspace{0.5mm}}c@{\hspace{0.5mm}}c@{\hspace{0.5mm}}c@{}}
% \includegraphics[width=0.16\linewidth]{samples/figure/main_comp/9179-001-input.png}&
% \includegraphics[width=0.16\linewidth]{samples/figure/main_comp/9179-001-input.png}&
% \includegraphics[width=0.16\linewidth]{samples/figure/main_comp/9179-001-sig.png}&
% \includegraphics[width=0.16\linewidth]{samples/figure/main_comp/9179-001-output.png}&
% \includegraphics[width=0.16\linewidth]{samples/figure/main_comp/9179-001-output.png}&
% \includegraphics[width=0.16\linewidth]{samples/figure/main_comp/9179-001-gt.png}&\\
% Input&Full-shadow face &shadow-free face &Mask &Shadowed portrait & Ground truth \\
% \end{tabular}
% \vspace{1mm}
% \caption{Illustration of our method. }
% \label{fig:main_comp}
% \vspace{-0.7em}
% \end{figure*}
\begin{figure*}[t]
\centering
\begin{tabular}{c@{\hspace{0.5mm}}c@{\hspace{0.5mm}}c@{\hspace{0.5mm}}c@{\hspace{0.5mm}}c@{\hspace{0.5mm}}c@{\hspace{0.5mm}}c@{}}
\includegraphics[width=0.19\linewidth]{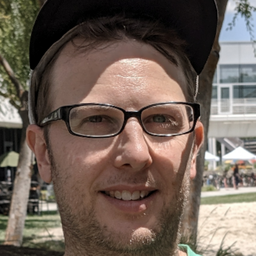}&
\includegraphics[width=0.19\linewidth]{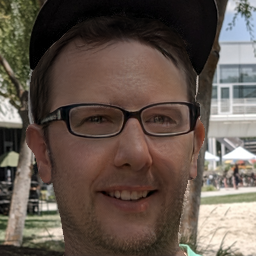}&
\includegraphics[width=0.19\linewidth]{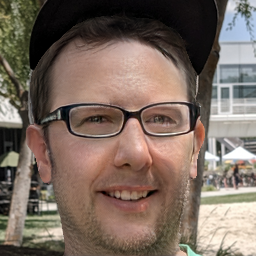}&
\includegraphics[width=0.19\linewidth]{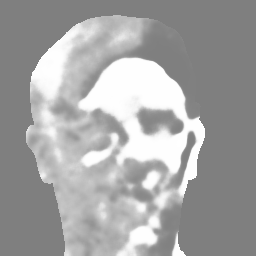}&
\includegraphics[width=0.19\linewidth]{samples/figure/more/first-in16-input.png}&\\
Input&Full-shadow portrait &Shadow-free portrait &Learned shadow mask &Recovered shadow portrait \\
\end{tabular}

\caption{Illustration of our method. We formulate the portrait shadow removal as an image decomposition problem. Our unsupervised method takes a single shadow portrait as input and can decompose it into a shadow-free portrait image, a full-shadow portrait image, and a shadow mask.}
\label{fig:illustration}
\vspace{-0.7em}
\end{figure*}

%% 3 
\section{Method}
% Problem formulation: layer decomposition. 
% Our framework: network architecture 
% Alternative optimization 
%% face segmentation 
%% regularization 
%% Loss functions 
% Discussion 
\label{method}
\subsection{Overview}
% problem formulation and overview 
We formulate the portrait shadow removal task as an image decomposition problem~\cite{zhang2020portrait}. To be specific, given an input shadowed portrait image $I$, we decompose it into $I_{free}$, $I_{full}$, and $M$:
\begin{align}
    I = I_{free} \odot M + I_{full} \odot (1-M), \label{formula:decomposition}
\end{align}
where $I_{free}$ and  $I_{full}$ denote shadow-free portrait and full-shadow portrait respectively. $M$ is a shadow mask which denotes the shadow region and intensity. $\odot$ is the Hadamard product. Decomposing a single shadowed portrait into $I_{free}$, $I_{full}$ and $M$ is a highly ill-posed problem. To achieve this, we leverage the generative priors to eliminate the layer ambiguity of the decomposition. 
To further constrain the decomposition process, we regularize the full shadow image $I_{full}$ using a color matrix $C \in [0,1]^{3 \times 3}$, 
\begin{align}
    C &= \left[
     \begin{matrix}
       \lambda_R & 0 & 0 \\
       0 & \lambda_G & 0 \\
       0 & 0 & \lambda_B
      \end{matrix}
      \right], \\
    I_{full}(x,y) &= C \times I_{free}(x,y),
\end{align}
where $x, y$ denote pixel positions and $\lambda$ denotes learnable shadow parameters for R, G and B channels. We will firstly introduce our network architecture at Section~\ref{method:architecture}, and then describe our proposed alternative optimization strategy at Section~\ref{method:optimization}. Finally, we introduce how to extend our framework to portrait tattoo removal and watermark removal tasks at Section~\ref{method:discussion}. 

\subsection{Network architecture}
\label{method:architecture}
% network architecture 
% free branch: stylegan good at generate high-quality faces. some other priors. GAN inversion. 
% full branch: sigmoid, initialization, 
% mask branch: noise, skip network, sigmoid, 
Following the image decomposition process defined in equation (\ref{formula:decomposition}), we design customized modules to learn $I_{free}$, $I_{full}$ and $M$ respectively, as illustrated in Figure~\ref{fig:overview}. Specifically, we use the pretrained StyleGAN2, denoted as $G$, as the branch for recovering underlying shadow-free portrait $I_{free}$, using optimization-based GAN inversion method~\cite{karras2020analyzing}. To reconstruct the full-shadow image $I_{full}$, we directly optimize a $3 \times 3$ color matrix through backpropagation. Note that although the same strategy can be applied to optimize our shadow mask $M$, we find it is prone to local minima during the mask $M$ optimization process. Thus, we adopt another small network \textit{MaskNet} $f_M$ and take a random initialized noise map $n_M$ as the input to the network to reconstruct the shadow mask $M$. We design the \textit{MaskNet} as an encoder-decoder network with skip connections, similar to Double-DIP~\cite{gandelsman2019double}. We apply a sigmoid function to MaskNet output to further regularize the value of learned mask to be in (0, 1). In summary, the StyleGAN2 latent code $w$, color matrix $C$ and the network parameters of \textit{MaskNet} $f_M$ need to be optimized. We design an effective progressive optimization strategy for this problem, as shown in Sec.~\ref{method:optimization}.

% \begin{algorithm}[h]
% \SetAlgoLined

% \KwResult{Write here the result }
%  initialization\;
% \SetKwInOut{hi}{hello} 
%  \While{While condition}{
%   instructions\;
%   \eIf{condition}{
%   instructions1\;
%   instructions2\;
%   }{
%   instructions3\;
%   }
%  }
%  \caption{Progressive optimization}
% \end{algorithm}

% \begin{algorithm}
% \caption{My algorithm}\label{euclid}
% \begin{algorithmic}[1]
% \Procedure{MyProcedure}{}
% \State $\textit{stringlen} \gets \text{length of }\textit{string}$
% \State $i \gets \textit{patlen}$
% \BState \emph{top}:
% \If {$i > \textit{stringlen}$} \Return false
% \EndIf
% \State $j \gets \textit{patlen}$
% \BState \emph{loop}:
% \If {$\textit{string}(i) = \textit{path}(j)$}
% \State $j \gets j-1$.
% \State $i \gets i-1$.
% \State \textbf{goto} \emph{loop}.
% \State \textbf{close};
% \EndIf
% \State $i \gets i+\max(\textit{delta}_1(\textit{string}(i)),\textit{delta}_2(j))$.
% \State \textbf{goto} \emph{top}.
% \EndProcedure
% \end{algorithmic}
% \end{algorithm}

\begin{algorithm}
    
  \caption{Progressive optimization}
  \label{algo:progressive}
  \begin{algorithmic}[1]
  
    % \State initialization
    % \State $\var{ini} = 4$
    % \State $\var{init} = 0$ 
    \Input{Shadowed portrait $I$, face parsing map $S$}
    \Output{Shadow-free portrait $I_{free}$, full-shadow         portrait $I_{full}$, blending mask $M$} 
    \phase{Initial face optimization.}
    % \Procedure {DriftingPhase}{}
      \\Sample 500 $\{z_i\}_{i=1}^{500}$ from Gaussian distribution; \\
      Infer $w$ space latents $\{w_i\}_{i=1}^{500}$ using $\{z_i\}_{i=1}^{500}$;\\
      Select $w_b$ which minimizes $\mathcal{L}_{LPIPS}(I, G(w_i))$;\\
      $w_b^{0}=w_b$;
      \For{$k=1$ to $K$}
        \State $I_{free}^{init}=G(w_b^{k-1})$;
        \State Loss=$\mathcal{L}_{LPIPS}(I_{free}^{init}, I)$;
        \State Update $w_b^{k-1}$ using ADAM algorithm;
        
      \EndFor \textbf{end for}
      \State \Return $I_{free}^{init}=G(w_b^{K})$.
    % \EndProcedure
    \phase{Color matrix and shadow mask optimization.}
    % \Procedure{FindingRoute phase}{}
      \\ Randomly initialize the MaskNet $f_M^0$ and the noise map $n_M$; \\$M^0 = f_M^0(n_M)$; \\
      Initialize diagonal element of color matrix to 0.5 to obtain $C^0$;
      \For{$p=1$ to $P$}
        \State $M^p = f_M^p(n_M)$;
        \State $\hat{I} = I_{free}^{init} \odot M^{p-1} + (C^{p-1} I_{free}^{init})\odot (1-M^{p-1})$;
        \State Loss=$MSE(\hat{I}, I)$;
        \State Update $C^{p-1}$ and $f_M^p$ using ADAM algorithm;
      \EndFor \textbf{end for}
      \State \Return $M^P$ and $C^P$.
    % \EndProcedure
    \phase{Facial features refinement.}
    % \Procedure{Facial features refinement.}{}
    \\ Initialization: $C^0=C^P, w^0=w_b^K,          I^0_{free}=I_{free}^{init}$;
      \For{$q=1$ to $Q$}
        \State $I^{q-1}_{free}=G(w^{q-1})$;
        \State  $\hat{I} = I_{free}^{q-1} \odot M^{P} + (C^{q-1} I_{free}^{q-1} )\odot (1-M^{P})$;
        \State Loss = $\mathcal{L}_{feat}(\hat{I}, I) + \mathcal{L}_{LPIPS}(\hat{I}, I)$;
        \State Update $C^{q-1}$ and $w^{q-1}$ using ADAM algorithm;
      \EndFor \textbf{end for}
      \\ $I_{free} = G(w^Q)$, $C=C^Q$, $M=M^P$, $I_{full}=C\times I_{free}  $;
      \State \Return $I_{free}$, $I_{full}$ and $M$.
    % \EndProcedure
  \end{algorithmic}
\end{algorithm}

\subsection{Progressive optimization}
\label{method:optimization}
\begin{figure*}[t]
\centering
\begin{tabular}{c@{\hspace{0.9mm}}c@{\hspace{0.7mm}}c@{\hspace{0.7mm}}c@{\hspace{0.7mm}}c@{\hspace{0.7mm}}c@{\hspace{0.7mm}}c@{}}
\rotatebox{90}{\small \hspace{18mm}}&
\includegraphics[width=0.19\linewidth]{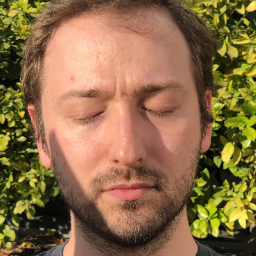}&
\includegraphics[width=0.19\linewidth]{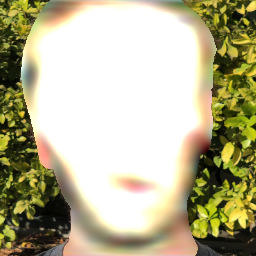}&
\includegraphics[width=0.19\linewidth]{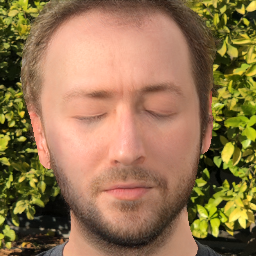}&
\includegraphics[width=0.19\linewidth]{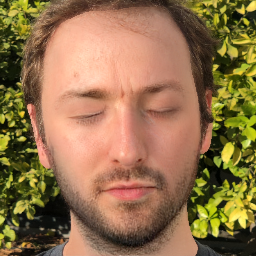}&
\includegraphics[width=0.19\linewidth]{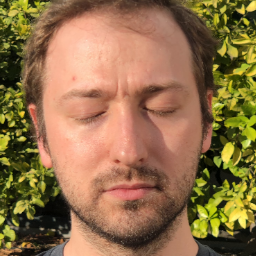}&\\
\rotatebox{90}{\small \hspace{25mm}}&
\includegraphics[width=0.19\linewidth]{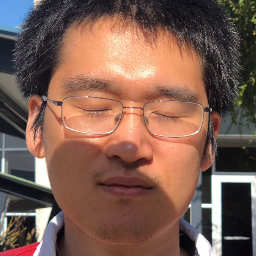}&
\includegraphics[width=0.19\linewidth]{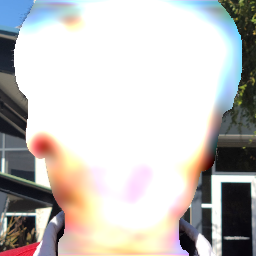}&
\includegraphics[width=0.19\linewidth]{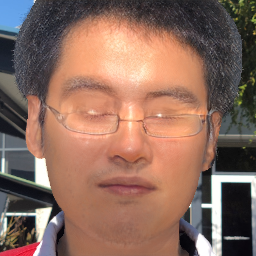}&
\includegraphics[width=0.19\linewidth]{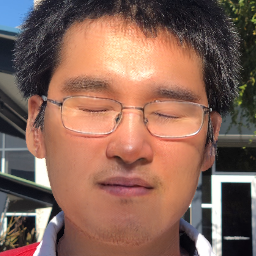}&
\includegraphics[width=0.19\linewidth]{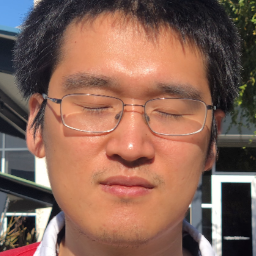}&\\
&Input &Our model, No GP  & Our model, JO & Our model & Ground truth \\
\end{tabular}
\vspace{0mm}
\caption{Controlled Experiments Results. Optimizing our framework without pretrained generative priors (No GP), fails at producing meaningful results. Jointly optimizing (JO) our framework will lose important perceptual facial details. Our progressive optimization strategy can effectively leverage the pretrained generative priors thus produce high-quality shadow-free portrait image.  }
\label{fig:ablation}
\vspace{-0.7em}
\end{figure*}

% regularization 
% face segmentation 
% loss functions 
% stage1: face segmentation; better sampling strategy; loss function 
% stage 3: face organ loss 
To produce high-quality shadow removal results, we need to carefully design the optimization process to utilize the GAN priors effectively. Joint optimization produces unpleasing artifacts, as shown in Figure~\ref{fig:ablation}. We instead adopt a progressive optimization strategy to guide the image recovering process step by step. We divide the optimization process into three stages, as explained in Algorithm~\ref{algo:progressive}.
\paragraph{Stage 1. Initial face optimization} We firstly use GAN inversion to project the shadowed image into the StyleGAN2 latent space for $K$ steps to obtain an initial shadow-free face. To avoid the disturbance of portrait background for GAN inversion quality, we only aim at recovering high-quality human faces through segmenting face parts with a pretrained face parsing model~\cite{yu2018bisenet}. We use LPIPS loss~\cite{Zhang_2018_CVPR} as the optimization goal for GAN inversion. 
% \begin{align}
%     \mathcal{L}_{LPIPS} = \Phi(I_{free}^{init} \odot S, I \odot S),
% \end{align}
\begin{align}
    \mathcal{L}_{LPIPS} = ||\Phi(I_{free}^{init} \odot S) - \Phi(I \odot S)||_2,
\end{align}
where $S$ is the segmentation mask for face obtained by~\cite{yu2018bisenet}, and $\Phi$ is the pretrained VGG-19 network~\cite{simonyan2014very,Zhang_2018_CVPR,Chen2017, lei2020blind}.
% , following the same setting as~\cite{karras2020analyzing}. 
We also add a regularization term to the StyleGAN2 noise map $n^G$ such that the image will not be projected into the noise latent space, as shown in~\cite{karras2020analyzing}. 
\begin{align}
    \mathcal{L}_{reg} &= \sum_{i, j} \mathcal{L}^{i, j}_{reg}.
\end{align}
where $\mathcal{L}^{i, j}_{reg}$ is defined by 
\begin{align}
    \mathcal{L}^{i, j}_{reg} &=\left(\frac{1}{r_{i, j}^{2}} \cdot \sum_{x, y} \boldsymbol{n}^G_{i, j}(x, y) \cdot \boldsymbol{n}^G_{i, j}(x-1, y)\right)^{2} \\ \notag
    &+\left(\frac{1}{r_{i, j}^{2}} \cdot \sum_{x, y} \boldsymbol{n}^G_{i, j}(x, y) \cdot \boldsymbol{n}^G_{i, j}(x, y-1)\right)^{2}, 
\end{align}
where $n^G_{i,j}$ is the $2^j$ times downsampled map of $i$-th noise map, $r_{i,j}$ denotes the resolution of $n^G_{i,j}$ and $x, y$ are pixel positions. 

In total, our loss function for \textit{Stage 1} can be formulated as 
\begin{align}
    \mathcal{L}_{S_1} = \mathcal{L}_{LPIPS} + \alpha \mathcal{L}_{reg}
\end{align}
Following~\cite{karras2020analyzing}, we set $\alpha$ to $10^5$ in our experiments. 

Latent Initialization: In order to ease the optimization process of this stage, we find it helpful to start with a good latent vector $w$ which approximates the face image $I \odot S$ well. Thus, instead of randomly initializing a latent vector from the prior distribution, we randomly sample 500 latent vectors fed into the pretrained StyleGAN2 to generate 500 images. Then we use $\mathcal{L}_{LPIPS}$ to select the best initial value for $w$. We empirically set $K$ to 300 to allow the network to produce the best shadow-free face approximation and to prevent fitting on portrait shadows. 

\paragraph{Stage 2. Color matrix and shadow mask optimization} After obtaining an approximate shadow-free face in \textit{Stage 1}, we aim at recovering a full-shadow portrait $I_{full}$ with the color matrix $C$, and estimating the shadow mask $M$ for image blending, following the definition in Equation (\ref{formula:decomposition}). In this stage, we fix the latent space of the StyleGAN2, and jointly optimize color matrix $C$ and parameters $\theta$ of MaskNet $f_M$ to minimize the reconstruction loss, which is defined by the $\mathcal{L}_2$ distance between reconstructed shadow portrait $\hat{I}$ and input shadowed portrait $I$ which served as ground truth:  
\begin{align}
    \min_{C, \theta}||\hat{I}_{C, \theta} - I||^2_2. 
    \label{formula:stage2}
\end{align}
In practice, we optimize objective function (\ref{formula:stage2}) for only 50 steps to guide the mask $M$ to learn the blending relationship instead of compensating for the face details. Please note that although there exists some face detail mismatch after the optimization of \textit{Stage 1}, the lighting appearance of shadow-free face $I_{free}$ is adequate to learn high-quality shadow mask $M$, as shown in Figure~\ref{fig:illustration}.

\begin{figure*}[t]
\centering
\begin{tabular}{c@{\hspace{0.5mm}}c@{\hspace{0.5mm}}c@{\hspace{0.5mm}}c@{\hspace{0.5mm}}c@{\hspace{0.5mm}}c@{\hspace{0.5mm}}c@{\hspace{0.5mm}}c@{}}
\includegraphics[width=0.135\linewidth]{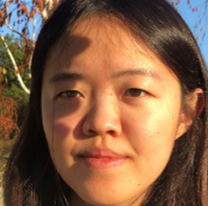}&
\includegraphics[width=0.135\linewidth]{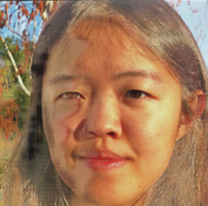}&
\includegraphics[width=0.135\linewidth]{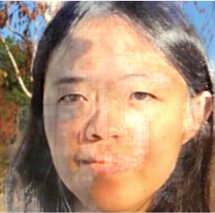}&
\includegraphics[width=0.135\linewidth]{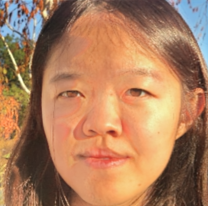}&
\includegraphics[width=0.135\linewidth]{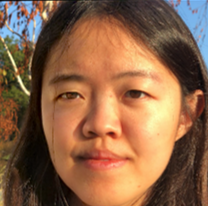}&
\includegraphics[width=0.135\linewidth]{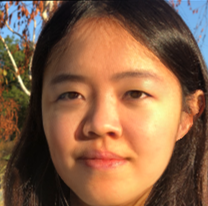}&
\includegraphics[width=0.135\linewidth]{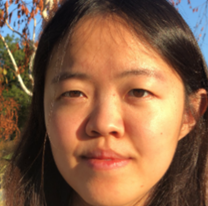}&\\
% \rotatebox{90}{\small \hspace{25mm}}&
\includegraphics[width=0.135\linewidth]{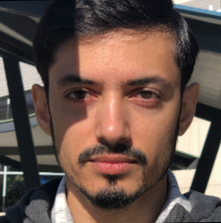}&
\includegraphics[width=0.135\linewidth]{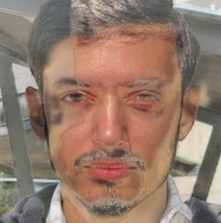}&
\includegraphics[width=0.135\linewidth]{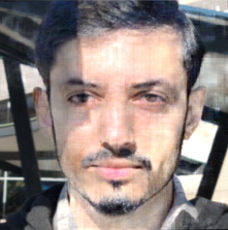}&
\includegraphics[width=0.135\linewidth]{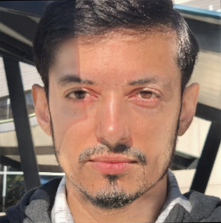}&
\includegraphics[width=0.135\linewidth]{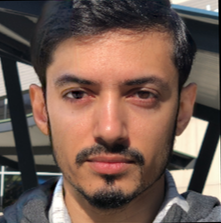}&
\includegraphics[width=0.135\linewidth]{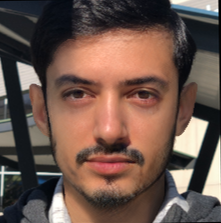}&
\includegraphics[width=0.135\linewidth]{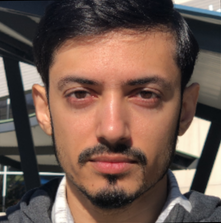}&\\
Input &ST-CGAN~\cite{wang2018stacked}&DAD~\cite{zou2020deep}&DHAN~\cite{cun2020towards}  &PSM~\cite{zhang2020portrait} & Ours & Ground truth \\
\end{tabular}
\vspace{0mm}
\caption{Comparison with baselines. Supervised shadow removal methods ST-CGAN, DAD and DHAN fail at shadow detection or shadow pixel relit, or both. Our unsupervised method can achieve comparable visual performance with state-of-the-art supervised portrait shadow removal method PSM~\cite{zhang2020portrait}.}
\label{fig:main_comp}
\vspace{-0.7em}
\end{figure*}

\paragraph{Stage 3. Facial features refinement} After the previous two optimization stages, we can now coarsely decompose $I$ into $I_{free}$ and $I_{full}$. However, the face reconstruction results of the first stage may miss perceptually important face details, such as eyeball colors and nose size,  since the optimization process is relatively short. Therefore, in this stage, we further improve the projection quality to the StyleGAN2 latent space to refine face details. Besides global perceptual loss applied on the whole face, we also propose facial feature loss to precisely optimize important face components:
\begin{align*}
     F=\{\text{nose}, \text{eyebrow}, \text{eyeball}, \text{mouth}, \text{glasses}\}. 
\end{align*}
We use the same parse map extracted in \textit{Stage 1} to identify important facial features. The LPIPS losses for face detail refinement are defined as 
\begin{align}
    \mathcal{L}_{feat} = \sum_{f \in F} \lambda_{\text {f}}  \Phi(f, \hat{f}),  
\end{align}
where $\text {f}$ is the element defined in $F$. 

We also optimize color matrix $C$ in this stage since the change of face detail may influence the estimation of full shadow images. We optimize this stage for 450 iterations, and the total optimization objective is 
% \begin{align}
%     \mathcal{L}_{S_3} = \mathcal{L}_{feat} + \mathcal{L}_{C}. 
% \end{align}
\begin{align}
    \min_{C, w} \mathcal{L}_{feat} + \mathcal{L}_{LPIPS}.
\end{align}

After optimization process, the generator of StyleGAN2 outputs a shadow-free portrait image $I_{free}$ which is served as final output result. 
\subsection{Extensions}
\label{method:discussion}
% general, can be used for other applications 
Our framework can also be extended to face tattoo removal and face watermark removal with only minimal modification. Note that this characteristic indicates better versatility of our framework than any existing supervised shadow manipulation methods. To demonstrate this potential, we synthesize faces with tattoos and watermarks respectively and adopt a similar optimization strategy for this layer decomposition problem. In these two tasks, we modify the problem formulation of equation (\ref{formula:decomposition}) into:
\begin{align}
    I = I_{clean} \odot M + I_{pure} \odot (1-M), \label{formula:tattoo}
\end{align}
where $I_{clean}$ and $I_{pure}$ denote the tattoo-free or watermark-free face and the pure-color face. Here we restrict the mask to be binary for better performance. 
\begin{align}
    \mathcal{L}_{binary} = min(|M-\mathbf{0}|, |M-\mathbf{1}|). 
\end{align}

\section{Experiments}
% Implementation details 
% Baselines 
% Ablation study 
% Quantitative results
% Qualitative results 
\subsection{Experimental setup}
\paragraph{Datasets.} 
% real-world dataset 
We evaluate our method on a real-world portrait shadow removal dataset which was proposed by ~\cite{zhang2020portrait}. The portrait shadow removal dataset contains 9 subjects and 100 shadowed portrait images in varied poses, shadow shapes, illumination conditions and shadow types which provides a challenging setting for portrait shadow removal task. For tattoo and watermark removal, we synthesize our own data based on CelebA-HQ dataset~\cite{CelebAMask-HQ}. 

\paragraph{Implementation details.}
We implement our method with PyTorch and conduct experiments on the NVIDIA RTX 2080Ti GPU. 
% We choose StyleGAN2 pretrained on 256$\times$256 FFHQ~\cite{karras2019style} high-quality face images as the generative model. 
We use the StyleGAN2 model which is pretrained on FFHQ~\cite{karras2019style} high-quality face images with resolution 256 $\times$ 256.
% We set the learning rate to 0.01 when optimizing StyleGAN2 latent space in \textit{Stage 1} and \textit{Stage 3}, fixed generator weights when optimizing color matrix and mask, and relax the generator in stage 3.
We set the learning rate to 0.01, 0.001 and 0.01 to  project shadow images into StyleGAN2 latent space, optimize MaskNet, and optimize color matrix respectively. 
% iteration of 3 stages 
% lr, loss weights, noise sampled from the uniform distribution. Following [], we perturb z randomly at each iteration to stablize the optimization process. 

\begin{figure*}[t]
    \begin{tabular}{c@{\hspace{0.5mm}}c@{\hspace{0.6mm}}c@{\hspace{0.6mm}}c@{\hspace{0.6mm}}c@{\hspace{0.6mm}}c@{}}
    \rotatebox{90}{\small \hspace{17mm} Input}&
    \includegraphics[width=0.24\linewidth]{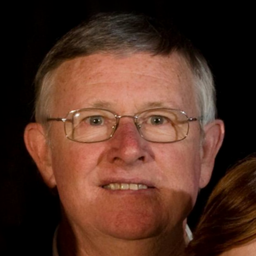}&
    \includegraphics[width=0.24\linewidth]{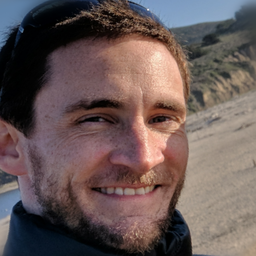}&
    \includegraphics[width=0.24\linewidth]{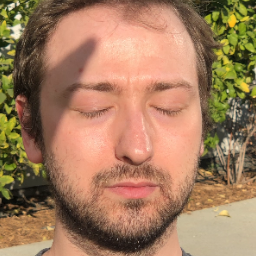}&
    \includegraphics[width=0.24\linewidth]{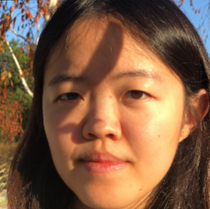}&\\
    \rotatebox{90}{\small \hspace{13mm} Our results}&
    \includegraphics[width=0.24\linewidth]{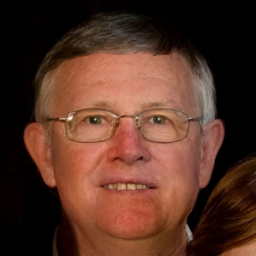}&
    \includegraphics[width=0.24\linewidth]{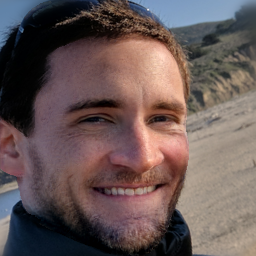}&
    \includegraphics[width=0.24\linewidth]{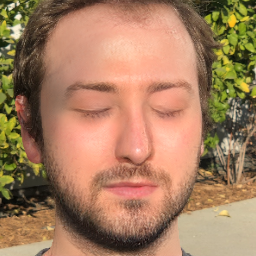}&
    \includegraphics[width=0.24\linewidth]{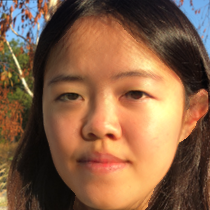}&\\
    \rotatebox{90}{\small \hspace{7mm} Learned shadow mask}&
    \includegraphics[width=0.24\linewidth]{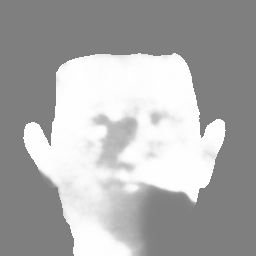}&
    \includegraphics[width=0.24\linewidth]{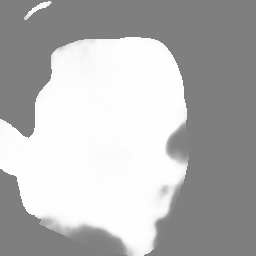}&
    \includegraphics[width=0.24\linewidth]{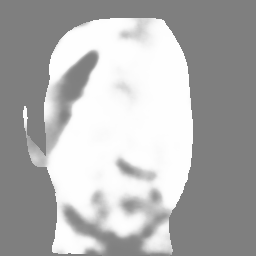}&
    \includegraphics[width=0.24\linewidth]{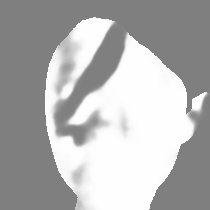}&\\
    \end{tabular}
    \vspace{-1mm}
    \caption{More Results. The bottom row shows the learned shadow mask. }
  \label{fig:more}
\end{figure*}

\begin{figure*}[t]
    \begin{tabular}{c@{\hspace{0.5mm}}c@{\hspace{0.6mm}}c@{\hspace{0.6mm}}c@{\hspace{0.6mm}}c@{\hspace{0.6mm}}c@{}}
    \rotatebox{90}{\small \hspace{17mm} Input}&
    \includegraphics[width=0.23\linewidth]{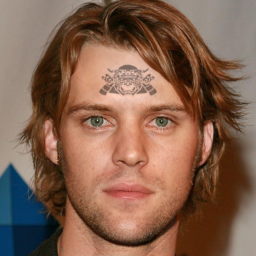}&
    \includegraphics[width=0.23\linewidth]{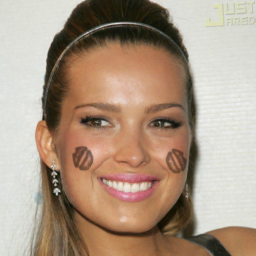}&
    \includegraphics[width=0.23\linewidth]{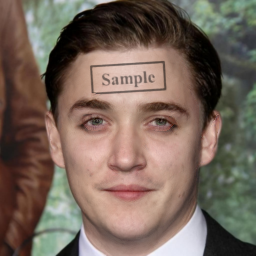}&
    \includegraphics[width=0.23\linewidth]{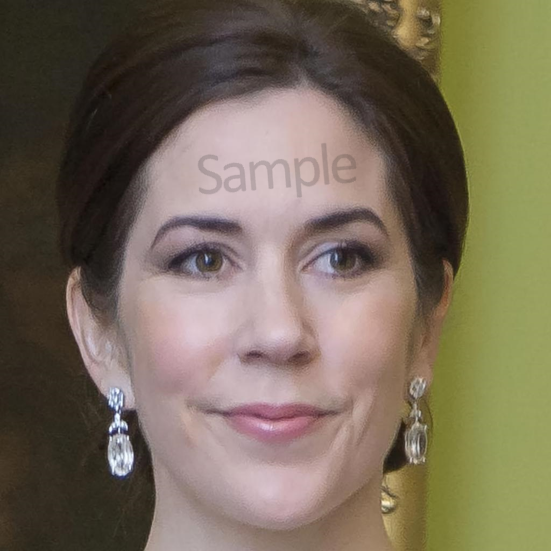}&\\
    \rotatebox{90}{\small \hspace{13mm} Our results}&
    \includegraphics[width=0.23\linewidth]{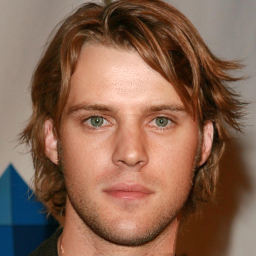}&
    \includegraphics[width=0.23\linewidth]{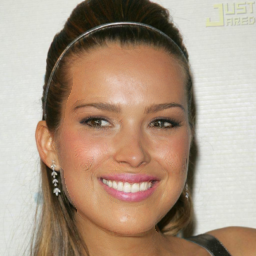}&
    \includegraphics[width=0.23\linewidth]{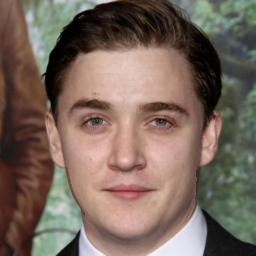}&
    \includegraphics[width=0.23\linewidth]{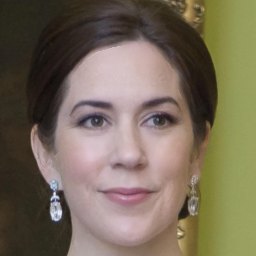}&\\
    \rotatebox{90}{\small \hspace{11mm} Learned mask}&
    \includegraphics[width=0.23\linewidth]{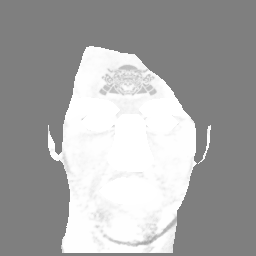}&
    \includegraphics[width=0.23\linewidth]{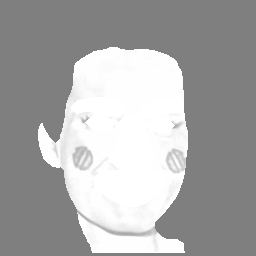}&
    \includegraphics[width=0.23\linewidth]{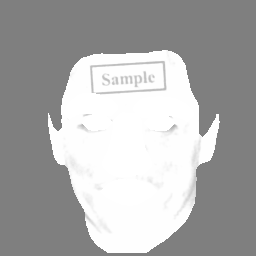}&
    \includegraphics[width=0.23\linewidth]{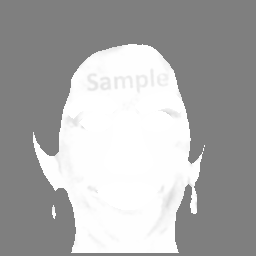}&\\ 
    &Tattoo removal&Tattoo removal&Watermark removal&Watermark removal&\\
    \end{tabular}
    \vspace{-1mm}
    \caption{Our facial tattoo and watermark removal results. Our method can recover high-quality clean portrait without any training data. The bottom row shows our learned blending masks. }
  \label{fig:tattoo}
\end{figure*}

% \begin{table}[]
% \setlength{\tabcolsep}{3mm}
% \begin{tabular}{l@{\hspace{2mm}}l@{\hspace{3mm}}c@{\hspace{3mm}}c@{\hspace{3mm}}c@{\hspace{3mm}}}
% \toprule
% & Methods & PSNR~$\uparrow$ & SSIM~$\uparrow$ & LPIPS~$\downarrow$ \\ \hline
% \multirow{4}{*}{\begin{tabular}[c]{@{}c@{}}Supervised \\ methods\end{tabular}}   & ST-CGAN~\cite{wang2018stacked}      & 14.25    &      0.512&  0.3031     \\
% & DAD~\cite{zou2020deep}          & 15.07     & 0.603     & 0.3225    \\
%  & DHAN~\cite{cun2020towards}         & 18.03     & 0.629     &  0.1607     \\
% & PSM~\cite{zhang2020portrait}          &  25.12    &  0.859     &   0.0874    \\ \hline
% \multirow{4}{*}{\begin{tabular}[c]{@{}c@{}}Unsupervised \\ methods\end{tabular}} 
% & DGP~\cite{pan2020exploiting}       &  N/A     &   N/A   &   N/A    \\
%  & Ours (No GP) &  13.88    & 0.707     &      0.3270 \\
%  & Ours (JO)    &  21.01    &  0.811    &  0.1288     \\
% & Ours & \textbf{21.64} & \textbf{0.820} & 0.1162     \\ 
% \bottomrule
% \end{tabular}
% \vspace{1mm}
% \caption{Quantitative evaluation of our model and baselines on portrait shadow removal.}
% \label{table:quant_other}
% \end{table}

\begin{table}[]
\setlength{\tabcolsep}{3mm}
\caption{Quantitative evaluation of our model and baselines on portrait shadow removal.}
\begin{tabular}{l@{\hspace{5mm}}l@{\hspace{7mm}}c@{\hspace{7mm}}c@{\hspace{5mm}}}
\toprule
& Methods  & SSIM~$\uparrow$ & LPIPS~$\downarrow$ \\ \hline
\multirow{4}{*}{\begin{tabular}[c]{@{}c@{}}Supervised \\ methods\end{tabular}}   & ST-CGAN~\cite{wang2018stacked}      &      0.512&  0.3031     \\
& DAD~\cite{zou2020deep}               & 0.603     & 0.3225    \\
 & DHAN~\cite{cun2020towards}          & 0.629     &  0.1607     \\
& PSM~\cite{zhang2020portrait}            &  \textbf{0.859}     &   \textbf{0.0874}    \\ \hline
\multirow{4}{*}{\begin{tabular}[c]{@{}c@{}}Unsupervised \\ methods\end{tabular}} 
& DGP~\cite{pan2020exploiting}         &   N/A   &   N/A    \\
 & Ours (No GP)     & 0.707     &      0.3270 \\
 & Ours (JO)       &  0.811    &  0.1288     \\
& Ours  & \textbf{0.820} & \textbf{0.1162}     \\ 
\bottomrule
\end{tabular}
\vspace{1mm}

\label{table:quant_other}
\end{table}

\vspace{-2mm}
\subsection{Baselines and controlled experiments}
% baselines: Sig20, Cun 
% ablations: joint optimization, w/o face segmentation

\textbf{Baselines:} Since our method is the first unsupervised portrait shadow removal framework, we can only compare our method with a set of state-of-the-art supervised shadow removal methods ST-CGAN~\cite{wang2018stacked}, DAD~\cite{zou2020deep}, and DHAN~\cite{cun2020towards} and portrait shadow removal methods PSM~\cite{zhang2020portrait} to prove our effectiveness. Our quantitative and qualitative results show that 
our method achieves superior performance than a set of general shadow removal methods and comparable results with the state-of-the-art portrait shadow removal approach.

% \begin{itemize}
%     \item ST-CGAN~\cite{wang2018stacked}, DAD~\cite{zou2020deep} and DHAN~\cite{cun2020towards} are three supervised learning methods training on large-scale shadow removal datasets. We use their provided pretrained model to evaluate their performance on shadowed portrait images.
%     \item PSM~\cite{zhang2020portrait} is a supervised-learning method for portrait shadow removal. They evaluate their method on both synthetic and real-world portrait shadow images. However, they did not publish their training/test codes or pretrained models. Thus, we directly use their public output results for comparisons. 

%     % \item SID~\cite{le2019shadow}
%     % \item DHAN~\cite{cun2020towards} utilizes a generative model for shadow detection and removal. Different from PSM~\cite{}, DHAN is a general-purpose shadow removal method. We diretly use the provided pretrained model to test on our dataset.
%     % \item PSM~\cite{zhang2020portrait} is a supervised-learning method for portrait shadow removal. They evaluate their method on both synthetic and real-world portrait shadow images. However, they did not publish their training/test codes or pretrained models. Thus, we directly use their public output results to compare with our method. 

% \end{itemize}

\textbf{Controlled experiments:} We also conduct several controlled experiments to evaluate the effectiveness of our proposed optimization strategies in Figure ~\ref{fig:ablation}. 

\textbf{(1) Joint optimization (JO):} Instead of optimizing different terms alternatively, we jointly optimize the shadow-free image, color matrix and shadow mask. 

\textbf{(2) No generative priors (NO GP):} We conduct another experiment without pretrained StyleGAN2 weights. We keep the same network architecture but randomly initialize the StyleGAN2 weights. 
%We show that this will lead to failed results due to the lack of abundant generative priors to guide the generation of shadow-free images. 

% \textbf{(3) Effectiveness of learned shadow mask and color matrix:} We conduct experiments to present how the leaned mask can boost the recovery of a shadow-free portrait image. Due to limited space please refer to supplementary materials for detailed results.

% \begin{itemize}
%     \item "Joint optimization (JO)". Instead of optimizing different terms alternatively, we jointly optimize the shadow-free image, color matrix and shadow mask. 
%     \item "No generative prior (NO GP)". We conduct another experiment without pretrained StyleGAN2 weights. We keep the same network architecture but randomly initialize the StyleGAN2 weights. We show that this will lead to failed results due to the lack of abundant generative priors to guide the generation of shadow-free images. 
% \end{itemize}

\vspace{-2mm}
\subsection{Quantitative results}
We use SSIM and LPIPS~\cite{Zhang_2018_CVPR} to quantitatively evaluate our method, since these two metrics mostly reflect perceptual qualities. The results are shown in Table~\ref{table:quant_other}. Compared with a series of supervised learning-based method~\cite{wang2018stacked,zou2020deep,cun2020towards}, our unsupervised method can achieve superior results in terms of all these evaluation metrics. This is because our method leverages rich generative facial priors which serve as important guidance to recover the underlying shadow-free portraits. Moreover, the results reflect that the supervised baseline methods own poor generalization ability to other domains such as portraits. Our method, however, is free of generalization issues thanks to its unsupervised nature. 
Compared with state-of-the-art supervised portrait shadow removal method~\cite{zhang2020portrait}, our method can achieve comparable performance. Moreover, as shown in Section~\ref{method:discussion} and Figure~\ref{fig:tattoo}, our framework can also be used as other portrait images decomposition tasks such as tattoo removal and watermark removal, which cannot be achieved by~\cite{zhang2020portrait}.

% \begin{table}
%   \caption{Frequency of Special Characters}
%   \label{tab:freq}
%   \begin{tabular}{ccl}
%     \toprule
%     Non-English or Math&Frequency&Comments\\
%     \midrule
%     \O & 1 in 1,000& For Swedish names\\
%     $\pi$ & 1 in 5& Common in math\\
%     \$ & 4 in 5 & Used in business\\
%     $\Psi^2_1$ & 1 in 40,000& Unexplained usage\\
%   \bottomrule
% \end{tabular}
% \end{table}

% \begin{table}[ht]
% \centering
% \setlength{\tabcolsep}{3mm}
% \begin{tabular}{l@{\hspace{9mm}}c@{\hspace{8mm}}c@{\hspace{8mm}}c@{\hspace{8mm}}}
% \toprule

%  & PSNR~$\uparrow$  & SSIM~$\uparrow$  & LPIPS~$\downarrow$  \\
% \midrule
% ST-CGAN~\cite{wang2018stacked} & 18.03 & 0.629 & 0.1248\\
% SID~\cite{le2019shadow} & 18.03 & 0.629 & 0.1248\\
% DHAN~\cite{cun2020towards} & 18.03 & 0.629 & 0.1248\\
% PSM~\cite{zhang2020portrait} & 25.12 & 0.859 & 0.0569\\
% \midrule
% DGP~\cite{pan2020exploiting} & N/A & N/A & N/A\\
% Ours (JO) & - &- & -  \\
% Ours (No GP) & - & - & \textbf{-}\\
% Ours & \textbf{21.64} & \textbf{0.820} & 0.0833 \\
% \bottomrule
% \end{tabular}
% \vspace{1mm}
% \caption{Quantitative evaluation among our model and baselines.}
% \label{table:quant_other}
% \end{table}


\vspace{-3mm}
\subsection{Qualitative results}
We also conduct qualitative comparisons between our method and baselines, as illustrated in Figure~\ref{fig:main_comp}.
ST-CGAN~\cite{wang2018stacked} and DAD~\cite{zou2020deep} both produce severe artifacts on real-world shadow portraits. Although the shadowed regions can be lightened up, they suffer from unnatural color distortion problems. Besides, the original well-lit face regions also contain  distorted color patches. The results indicate that~\cite{wang2018stacked, zou2020deep} can neither accurately detect shadow regions nor relit shadowed pixels. The recovered shadow-free portrait of DHAN~\cite{cun2020towards} differs a lot from ground-truth portraits, especially at no-shadow regions. The results indicate that their method performs well at shadow detection but poor at shadow removal. PSM~\cite{zhang2020portrait} is a state-of-the-art supervised portrait shadow removal method that is trained on the same dataset as ours. Our method is free of any external training data with the aid of generative priors. Our method can achieve comparable visual results with PSM~\cite{zhang2020portrait} but own greater application scenarios (e.g., tattoo or watermark removal) than any existing supervised shadow removal methods. 

\vspace{-2mm}
\subsection{Limitations}
While our method can can achieve comparable performance with state-of-the-art supervised methods, we do observe some unpleasing artifacts. For example, when the portrait image contains fine-grained details such as wrinkles and bushy beards (see the first case in Fig.~\ref{fig:tattoo}), these details may be smoothed in the output image. This is due to the imperfect reconstruction results of GAN inversion, which is also an active research area. 
Besides, our method may not work well when the portrait accessories or clothing are not in the training set of StyleGAN2. Thus, more powerful expression ability and GAN inversion techniques of StyleGAN2 can be further explored to improve our face restoration quality. 
% (a) fine-grained facial details 
% (b) extreme pose
% While the quality of our framework is better than competing work and comparable with state-of-the-art supervised methods, we note that sometimes the fine-grained details of recovered facial images are smoothed, such as wrinkles and bushy beards. This is due to the imperfect GAN inversion, which is still an active research area. We will supplement our worst/failure cases in the final revision.

\vspace{-3mm}
\section{Conclusion}
We proposed the first unsupervised method for portrait shadow removal which needs only one input shadow portrait image. We have shown that the generative priors can be used in this unsupervised layer decomposition setting to handle unknown degradation processes which cannot be accomplished by existing GAN-inversion methods. Meanwhile, we designed progressive optimization techniques to guide the image decomposition and reconstruction process. Then, we achieved comparable performance with existing state-of-the-art supervised-based shadow removal methods, demonstrating the effectiveness of our unsupervised method. Finally, we have shown two extension applications (e.g., portrait tattoo removal and watermark removal) of our method to demonstrate that our method can 
serve as a unified framework for portrait image decomposition tasks.
% method using generative facial prior. Our method can recover high-quality underlying shadow-free portrait without any training data. We formulate the portrait shadow removal task as an image decomposition problem from the shadow generation nature. We carefully design the multi-branch optimization-based architecture to effectively leverage the abundant facial prior in pretrained StyleGAN. We propose an alternative optimization strategy for the image decomposition. We demonstrate the effectiveness of our method through extensive experiments. We also extend our method to portrait tattoo removal and landmark removal to prove the versatility of our framework. 

\begin{acks}
We thank Xuaner Zhang, Jiapeng Zhu and anonymous reviewers for helpful discussions on the paper. 

\end{acks}

\bibliographystyle{ACM-Reference-Format}
\bibliography{sample-base}


\end{document}